\documentclass[journal]{IEEEtran}
\usepackage{cite}
\usepackage{booktabs}
\usepackage{arydshln} 
\usepackage{multirow}
\usepackage[table]{xcolor}
\usepackage{amsmath}
\usepackage{multirow}
\usepackage{algorithm}
\usepackage{algorithmic}
  
\usepackage{array}

\usepackage[colorlinks,
linkcolor=green, 
anchorcolor=blue,
citecolor=blue]{hyperref}

\usepackage{color}            

\usepackage{mathrsfs}
\usepackage{graphicx}
\usepackage{subfigure}
\usepackage{float}
\usepackage{tabularx} 
\begin{document}

\title{Net2Net: When Un-trained Meets Pre-trained Networks for Robust Real-World Denoising}

\author{Weimin~Yuan,
        Yinuo~Wang,
        Cai~Meng

\thanks{W. Yuan, Y. Wang and C. Meng are with the Image Processing Center, Beihang
University, Beijing 100191, China (e-mail: yuanweimin@buaa.edu.cn; Tsai@buaa.edu.cn).}}

\maketitle

\begin{abstract}
    Traditional denoising methods for noise removal have largely relied on handcrafted priors, often perform well in controlled environments but struggle to address the complexity and variability of real noise. In contrast, deep learning-based approaches have gained prominence for learning noise characteristics from large datasets, but these methods frequently require extensive labeled data and may not generalize effectively across diverse noise types and imaging conditions. In this paper, we present an innovative method, termed as Net2Net, that combines the strengths of untrained and pre-trained networks to tackle the challenges of real-world noise removal. The innovation of Net2Net lies in its combination of unsupervised DIP and supervised pre-trained model DRUNet by regularization by denoising (RED). The untrained network adapts to the unique noise characteristics of each input image without requiring labeled data, while the pre-trained network leverages learned representations from large-scale datasets to deliver robust denoising performance. This hybrid framework enhances generalization across varying noise patterns and improves performance, particularly in scenarios with limited training data. Extensive experiments on benchmark datasets demonstrate the superiority of our method for real-world noise removal.
\end{abstract}


\IEEEpeerreviewmaketitle

\section{Introduction}

\IEEEPARstart{R}{eal} world denoising \cite{buades2005review,yuan2024simultaneous,elad2023image,izadi2023image,milanfar2024denoising,yuan2024weighted,yuan2023guided,li2025gs2pose}, as a practical scenario with high application value in image restoration, has attracted increased attention in medical imaging, surveillance imaging, satellite remote sensing, and traffic hub monitoring, and so on. Real-world denoising is an inherently ill-posed inverse problem because degradation processes are irreversible \cite{8026108,yuan2025image,wang2025mamba,yuan2025guided,wang2024samihs,li2024single,wang2023coam,meng2023oscillating,li2022review,yuan2021efficient,yuan2019sgm}. The vast variability in natural image content makes it essential to use prior information to regularize the solution space and obtain accurate estimations of the latent image. A variety of methods have been proposed in the past several decades, which could be divided into learning prior based and model prior based methods.

Denoising networks have been trained using synthetic noise models \cite{7839189,8365806}, such as Gaussian or Poisson noise, which are artificially added to images for training and evaluation purposes. However, real-world noise, influenced by various factors within the Image Signal Processing pipeline such as demosaicing and gamma correction, exhibits a distinct signal dependency and often follows distributions that differ from synthetic counterparts \cite{Ryou_2024_CVPR}. This divergence between synthetic and real-world noise distributions raises significant generalization issues when applying denoising models to real noisy images. While efforts have been devoted to creating datasets with clean and noisy images in the real-world \cite{7780555,8578280,8099777,xu1804real,8953965,lebrun2015noise,anaya2018renoir}, collecting these datasets poses a significant challenge. Although most of the existing denoising models have achieved remarkable success in a specific single scene, their performance is not satisfactory when facing unknown and diverse noise scenes.
Image denoisers trained with supervision on real noise datasets, however, come with their own set of flaws. Notably, even advanced denoisers \cite{8954448,9010324,9157160} encounter difficulty generalizing to variations in noise distributions, which arise from factors such as different camera sensor types, shooting environments, and ISP process. To this end, various self-supervised methods \cite{9577798,Wang_2022_CVPR,10203304,Pan_2023_ICCV} have been emerged as promising solutions for denoising, aiming to reduce the dependence on paired noisy-clean image datasets. Nevertheless, these methods often fall short in performance when compared to supervised counterparts.

On the other hand, representative model prior includes total variation (TV) \cite{xx040605412}, sparsity \cite{6814320} and low-rank \cite{6909762}. However, single physical prior based methods may not be sufficient for capturing the characteristics of degradation scenarios. Although promising performance has been achieved by these prior-based methods, the performance is not always desirable due to the inconsistency between the adopted prior and real environment. Moreover, these methods are shallow models which might be with limited capacity of handling complex data.

There are two important meanings of combining the unsupervised network DIP and the pre-trained model through RED. For the overfitting defect of the former, the latter can be used as a regularization term to effectively control the convergence of the algorithm; at the same time, the poor generalization defect of the latter trained based on simulation data in real noise removal is effectively improved by combining with the unsupervised network.
In short, by combining the unsupervised network and the pre-trained model trained based on simulation data, superior performance is achieved in real scenarios.

In summary, the main contributions of this paper are given as follows: 

\begin{itemize}
    \item We propose a novel denoising framework combining untrained and pre-trained networks: The Net2Net framework leverages the strengths of a Deep Image Prior (DIP) for unsupervised noise adaptation and a pre-trained network (DRUNet) for regularization. This hybrid approach addresses the challenges of real-world noise removal by capturing the specific noise characteristics of each image without requiring extensive labeled data while enhancing stability through pre-trained network regularization.
    
    \item Net2Net significantly enhances its ability to generalize to various real-world noise distributions by utilizing the unsupervised module to handle unknown noise patterns and the pre-trained module to ensure robust denoising. The framework is versatile and performs well not only on synthetic noise datasets but also in low-data environments with diverse imaging conditions. 
    \item Extensive experiments on several real-world noise datasets demonstrate the effectiveness and robustness of Net2Net in real-world denoising tasks.
\end{itemize}

The reminder of this paper is organized as follows. Section 2 introduces the related work. Section 3 presents the proposed method. Section 4 provides the analysis of the feasibility of our method, followed by experimental validation in Section 5. Finally, Section 7 concludes this paper.

\section{Related Works}
\subsection{Supervised Method}

There have been significant advancements in the area of supervised synthetic noise removal, such as Additive White Gaussian Noise (AWGN). The initial breakthroughs were largely driven by CNN-based models, pioneering work DnCNN \cite{7839189}, outperform traditional counterparts \cite{4378954,
6909762}, leading the stage for further innovations in this domain. To improve model flexibility, FFDNet \cite{8365806} trains a conditional non-blind denoiser with a manually adjusted noise-level map. By giving high-valued uniform maps to FFDNet, over-smoothed results may obtained in real noise removal. However, such above models do not generalize well on real-world applications due to domain discrepancy between realistic and synthetic noise, which has been known as \textit{simulated-to-real-gap} \cite{kohler2019toward}. Compared to pixel independent AWGN, real-world noises are much more complex, which can be spatially variant, spatially correlated, signal dependent, and even device dependent \cite{zhou2020awgn}.

To overcome this limitations, several approaches, such as CBDNet \cite{8954448}, PD \cite{zhou2020awgn}, RIDNet \cite{9010324}, AINDNet \cite{9157160}, InvDN \cite{9577357} train their models on realistic noisy-clean pairs.

To be specific, CBDNet \cite{8954448} address the problem of real-noise removal by modeling the realistic noise using the in-camera pipeline model \cite{liu2007automatic}. CBDNet also trains an explicit noise estimator and sets a larger penalty for under-estimated noise. But CBDNet still cannot fully characterize real noises. In \cite{zhou2020awgn}, Zhou et al. propose a novel approach to boost the performance of a real image denoiser which is trained only with AWGN. A deep model that consists of a noise estimator and a denoiser with mixed AWGN and Random Value Impulse Noise (RVIN) is trained. And then
investigate Pixel-shuffle Down-sampling (PD) strategy to adapt the trained model to real noises. RINDNet \cite{9010324} selectively learns distinctive distinguishing features through an attention mechanism. AINDNet \cite{9157160} uses a migration learning strategy to propose a denoising structure with a strong generalization.

In general, there are two main drawbacks of these approaches from the perspective of practical applications. One is the high dependency on the quality and the size of the training dataset. Collecting such image pairs is time-consuming and requires much of human efforts, especially when the labels needs deep domain knowledge such as medical or seismic images. The other drawback is the generalization ability of a trained network. When the noisy distribution is complicated and not contained in the training set, the results of the deep learning method can be deteriorated significantly, even worse than non-learning based methods.

\subsection{Self-Supervised Method}

Collecting clean and noisy pair data from the real-world is cost-intensive. In some certain scenarios, such as medical imaging and electron microscopy, constructing such datasets can be impractical or even infeasible. Therefore, there has been an active research on exploring self-supervised denoising models \cite{lehtinen18a,9577798,Wang_2022_CVPR,10203304,Pan_2023_ICCV} have relaxed the clean-noisy pairs dataset requirement. These approaches demonstrate the feasibility of training denoising networks using datasets
comprising solely noisy pairs or individual noisy images, without the need for a clean and noisy pair dataset. However, in real-world scenarios, noise often exhibits spatial correlation, which contradicts the pixel-wise independence noise assumption \cite{lehtinen18a}. 

Recent studies have presented self-supervised denoising methods specilized for real-world scenarios \cite{Neshatavar_2022_CVPR,9878719,Pan_2023_ICCV,chen2024exploring}. For example, AP-BSN \cite{9878719} utilizes pixel-shuffle downsampling (PD) to disrupt the noise correlation and employed asymmetric PD stride factors for training and inference.

\subsection{Un-Supervised Method}

By taking the structure of a randomly initialized network as an image prior, DIP \cite{Ulyanov2017DeepIP} shows that a deep CNN has an intrinsic ability to learn the uncorrupted image. DIP has an edge in real-world scenes because it needs no training data. Such advantages of DIP have attracted widespread attention, and motivated a large number of studies. For example, Deep decoder (DD) \cite{heckel2018deep} learns to decode noise tensor with no convolutions. By leveraging the CNN architecture search, neural architecture search deep image prior (NAS-DIP) \cite{NAS-DIP} and image-specific neural architecture search deep image prior (ISNAS-DIP) \cite{Arican2021ISNASDIPIN}
search for neural architectures that capture stronger structured priors, but the need for candidate comparison dramatically prolongs the
restoration time. However, NAS-based \cite{NAS-DIP,Arican2021ISNASDIPIN} are typically heavily
parameterized and prone to over-fitting. Frequency-band correspondence measure is introduced in \cite{54} to characterize the spectral bias of DIP, revealing that low-frequency image signals are learned more efficiently than high-frequency ones. DRP-DIP \cite{Li_2023_CVPR} freezes randomly initialized network weights and reducing the network depth to speed up DIP. These unsupervised approaches have proven effective in restoration tasks, achieving impressive results comparable to
supervised learning methods. However, they often struggle with handling highly complex or corrupted images due to their reliance on learned distributions and intrinsic image properties, which may not fully capture intricate details and show limited generalization to other restoration tasks.

\section{Preliminary}

Table \ref{table_symbols} summarizes the frequently-used notations used in this paper. We use the normal symbol for scalar and bold symbol for vector.

\begin{table*}[h]
\centering
\caption{Summary of variables and the descriptions.}
\label{table_symbols}
\begin{tabular}{llll}
\toprule
  \textbf{Notation}   & \textbf{Description}&\textbf{Notation}&\textbf{Description} \\ \midrule
  
$\boldsymbol{x}$, $\boldsymbol{y}$ & Clean, corrupted images &

$\boldsymbol{N}$ & Noise  \\

$\boldsymbol{\Theta}$ & Parameters of deep prior &

$\boldsymbol{z}$ &  Initialized noise
vector  \\

$\boldsymbol{U}(\cdot)$& Un-trained deep prior  &

$\boldsymbol{P}(\cdot)$ & Pre-trained deep prior \\

$\boldsymbol{u}$ & Lagrangian multiplier &
$\boldsymbol{P}_{BF}(\cdot)$ & Bias-free pre-trained deep prior  \\

$\boldsymbol{s}$, $\boldsymbol{w}$ &Auxiliary variables  & $\boldsymbol{B}$, $\boldsymbol{C}$ &  Lagrange multipliers  \\

$\tau_{1}$, $\tau_{2}$& Non-negative penalty factors & $\eta$ & Step factor  \\\bottomrule

\end{tabular}
\end{table*}

\subsection{Deep image prior (DIP)}
DIP \cite{Ulyanov2017DeepIP} leverages the intrinsic structure of CNNs (un-trained) to restore image without the need for a specific dataset of clean images. Unlike supervised learning methods that rely on large datasets to learn the mapping, DIP relies instead on the internal structure of CNN architecture itself to impose a strong prior on the space of natural images. DIP implicitly regularizes the process of restoration by removing the explicit regularization, and the core idea behind DIP is to minimize:
\begin{equation}\label{g1-5}
\begin{aligned}
    \hat{\boldsymbol{\Theta}}=\arg \min_{\boldsymbol{\Theta}}\|\boldsymbol {U}_{\boldsymbol{\Theta}}(\boldsymbol{z})-\boldsymbol {y}\|_{2}^{2}  \hspace{0.5cm} s.t. \hat{\boldsymbol{x}}=\boldsymbol {U}_{\hat{\boldsymbol{\Theta}}}(\boldsymbol{z})
\end{aligned}
\end{equation}
here, $\boldsymbol{\Theta}$ represents the parameters of CNN, $\boldsymbol{z}$ is a randomly initialized noise vector. The optimization process is performed through gradient descent method, iteratively updating $\boldsymbol{\Theta}$ for minimizing $\|\boldsymbol {U}_{\boldsymbol{\Theta}}(\boldsymbol{z})-\boldsymbol {y}\|_{2}^{2}$. Given the optimal  $\hat{\boldsymbol{\Theta}}$, recovering result $\hat{\boldsymbol{x}}$ is the output of network $\boldsymbol{U}_{\hat{\boldsymbol{\Theta}}}(\boldsymbol{z})$. The key advantage lies in without relying on specific training datasets, making DIP to be a valuable tool for image denoising, inpainting, and super-resolution \cite{heckel2018deep,NAS-DIP,Arican2021ISNASDIPIN,54,Li_2023_CVPR} with no access to clean data.

\subsection{Regularization by Denoising (RED)}
Compared to forthright employ denoiser into a regularization in plug-and-play (PNP) \cite{venkatakrishnan2013plug}, \textit{RED} \cite{doi:10.1137/16M1102884} is a radically new way to exploit an image denoising engine. \textit{RED} embeds an arbitrarily specified denoiser $\boldsymbol{P}(\boldsymbol{x})$ into the regularization for image restoration. The mathematical framework of \textit{RED} is the inner product between $\boldsymbol{x}$ and its denoising residual $\boldsymbol{x}-\boldsymbol{P}(\boldsymbol{x})$ to construct an explicit regularizer as the form:
\begin{equation}\label{red}
\begin{aligned}
\rho\boldsymbol{(x)}=\frac{1}{2}
\boldsymbol{x}^\mathrm{T}[\boldsymbol{x}-\boldsymbol{P}(\boldsymbol{x})]
\end{aligned}
\end{equation}
The advantage of using an explicit regularizer is that a wide variety of optimization algorithms (such as fixed-point, steepest-descent) can be used to solve Eq. (\ref{red}) and their convergence can be tractably analyzed under the following RED conditions on the denoiser $\boldsymbol{P}(\boldsymbol{x})$:

\textit{Definition 1(RED conditions \cite{doi:10.1137/16M1102884,cohen2021regularization})}: The \textit{RED conditions} for a denoiser $\boldsymbol{P}(\boldsymbol{x})$ are:
\begin{itemize}
    \item  \textit{Differentiability}: The denoiser $\boldsymbol{P}(\boldsymbol{x})$ is differentiable.
    \item  \textit{Homogeneity}: $\boldsymbol{P}[(1+\epsilon)\boldsymbol{x}]=(1+\epsilon)\boldsymbol{P}(\boldsymbol{x})$ for small positive $\epsilon$.
    \item \textit{Jacobian Symmetry}. The Jacobian matrix of $\boldsymbol{P}(\boldsymbol{x})$ meets symmetry: $J \boldsymbol{P}(\boldsymbol{x})=[J \boldsymbol{P}(\boldsymbol{x})_{}]^{T}$.
\end{itemize}

If a denoiser $\boldsymbol{P}(\boldsymbol{x})$ satisfies \textit{RED conditions}, a key and highly beneficial property is obtained for RED regularization, namely $\boldsymbol{P}(\boldsymbol{x})$ is a convex function and its gradient can be efficiently computed as the denoising residual:
\begin{equation}\label{red_add1}
\begin{aligned}
    \nabla \rho\boldsymbol{(x)}=\boldsymbol{x}-\boldsymbol{P}(\boldsymbol{x})
\end{aligned}
\end{equation}

\textit{RED} has demonstrated impressive recovery performance while providing global convergence guarantees. However, various common denoisers do not satisfy the \textit{RED conditions} \cite{Reehorst2018RegularizationBD}. In such cases, Eq. (\ref{red_add1}) does not hold and the corresponding convergence analysis is not hold.

\section{The Proposed Net2Net}\label{3}
\begin{figure*}[ht]
\centerline{\includegraphics[width=1\linewidth]{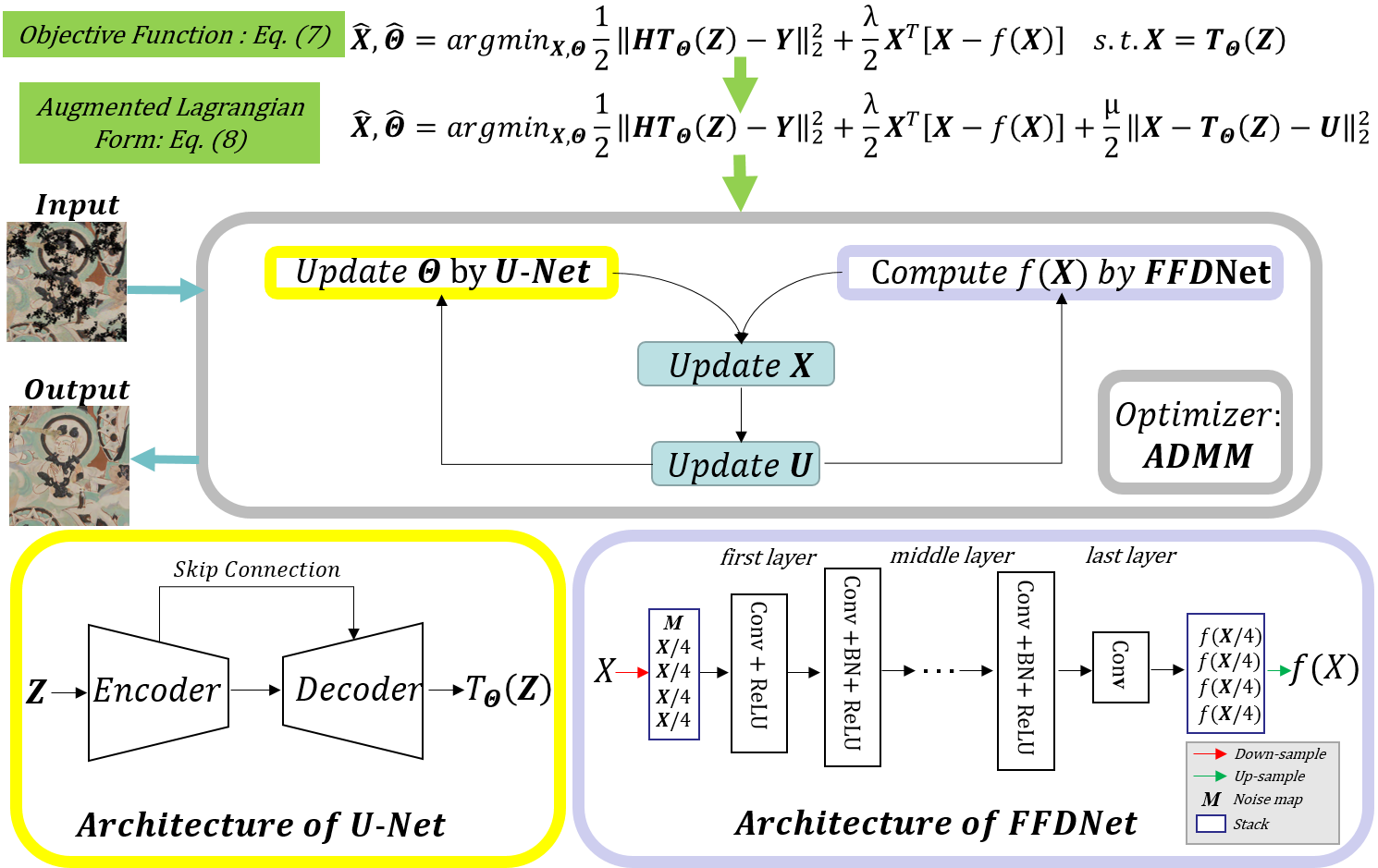}}
 \caption{The overall pipeline of Net2Net. In brief, Net2Net consists of two sub-networks, namely unsupervised and pre-trained networks and connected by RED.} 
\label{flow_net2net}
\end{figure*}

The flowchart of the proposed Net2Net for real noise removal is depicted in Fig. \ref{flow_net2net}. The subsequent section elaborate the details.

The noise in real-world scenes typically arises from multiple sources, including capturing devices, data transmission media, and image quantization. The complexity of this noise generation process makes it challenging to accurately characterize the noise and recover the underlying clean image. This challenge is central to real noise removal. From a practical perspective, there are two major limitations in existing deep learning-based denoising methods. First, their performance heavily relies on the quality and quantity of the training dataset. Collecting high-quality image pairs can be time-consuming and labor-intensive, especially for domains requiring specialized knowledge, such as medical imaging or seismic data. Second, the generalization ability of these networks is often limited. When the noise distribution in real images deviates from the training set, the performance of deep learning methods can degrade significantly, sometimes even falling behind traditional non-learning-based methods.

To address these challenges, we propose Net2Net, a novel denoising framework that integrates both an unsupervised denoising module (DIP) and a pre-trained network (DRUNet), leveraging the strengths of both approaches. By combining these modules with a regularization strategy, Net2Net effectively mitigates the need for large labeled datasets and improves generalization to complex noise distributions. To be specific, the objective function of our method is:
\begin{equation}\label{g2}
\begin{aligned}
    \hat{\boldsymbol {x}},\hat{\boldsymbol{\Theta}}=&\arg \min_{\boldsymbol{x,\Theta}}{\frac{1}{2}\|\boldsymbol{U}_{\boldsymbol{\Theta}}(\boldsymbol{z})-\boldsymbol {y}\|_{2}^{2}+ \frac{\lambda}{2}
\boldsymbol{x}^\mathrm{T}[\boldsymbol{x}-\boldsymbol{P}(\boldsymbol{x})]},\\& \quad s.t.\quad \boldsymbol{x}=\boldsymbol {U}_{\boldsymbol{\Theta}}(\boldsymbol{z})
\end{aligned}
\end{equation}
here, $\lambda$ controls the RED regularization strength.

\section{Optimizer}\label{3.2}
ADMM \cite{56} is employed in our method to decompose Eq. (\ref{g2}) into three sub-problems. We turn the constraint in Eq. (\ref{g2}) into a penalty using the Augmented Lagrangian:
\begin{equation}\label{g3}
\begin{aligned}
    \hat{\boldsymbol {x}},\hat{\boldsymbol{\Theta}}=&\arg \min_{\boldsymbol{x,\Theta}}{\frac{1}{2}\|\boldsymbol{U}_{\boldsymbol{\Theta}}(\boldsymbol{z})-\boldsymbol {y}\|_{2}^{2}+ \frac{\lambda}{2}
\boldsymbol{x}^\mathrm{T}[\boldsymbol{x}-\boldsymbol{P}(\boldsymbol{x})]}+\\&{\frac{\mu}{2}\|\boldsymbol{x}-\boldsymbol {U}_{\boldsymbol{\Theta}}(\boldsymbol{z})-\boldsymbol{u}\|_{2}^{2}}
\end{aligned}
\end{equation}
$\mu$ is a introduced ADMM factor, $\boldsymbol{u}$ is corresponding Lagrangian multiplier. The ADMM algorithm is used to addresses Eq. (\ref{g3}), and it transforms the problem into solving three sub-problems: $\boldsymbol{\Theta}$, $\boldsymbol{x}$ and $\boldsymbol{u}$ sequentially.

\subsection{Update of sub-problem}

When $\boldsymbol{x}$ and $\boldsymbol{u}$ are fixed, $\boldsymbol{x}^\mathrm{T}[\boldsymbol{x}-\boldsymbol{P}(\boldsymbol{x})]$ in Eq. (\ref{g3}) is omitted:
\begin{equation}\label{g4}
\begin{aligned}
    \hat{\boldsymbol{\Theta}}=\arg \min_{\boldsymbol{\Theta}}{\|\boldsymbol {U}_{\boldsymbol{\Theta}}(\boldsymbol{z})-\boldsymbol {y}\|_{2}^{2}}+{\mu\|\boldsymbol{x}-\boldsymbol {u}-\boldsymbol{U}_{\boldsymbol{\Theta}}(\boldsymbol{z})\|_{2}^{2}}
\end{aligned}
\end{equation}
Eq. (\ref{g4}) shows a resemblance with the optimization in DIP, which uses back-propagation. The second term forces $\boldsymbol{U}_{\boldsymbol{\Theta}}(\boldsymbol{z})$ to approach $(\boldsymbol{x}-\boldsymbol{u})$, which provides a stabilizing and robustifying effect. We optimize Eq. (\ref{g4}) by gradient descent. UNet \cite{ronneberger2015u} is employed in our framework as un-trained architecture   $\boldsymbol{U}_{\boldsymbol{\Theta}}(\cdot)$. Following \cite{Ulyanov2017DeepIP}, the network input $\boldsymbol{z}$ is given as uniform noise between 0 and 0.1 with a depth of 32 by fault.


\subsection{Update of sub-problem}\label{4.3.2}

By freezing $\boldsymbol{\Theta}$ and $\boldsymbol{U}$ in Eq. (\ref{g3}), we have:
\begin{equation}\label{g5-00001}
\begin{aligned}
    \hat{\boldsymbol {x}}=\arg \min_{\boldsymbol{x}}{\frac{\mu}{2}\|\boldsymbol{x}-\boldsymbol{C}\|_{2}^{2}}+\frac{\lambda}{2}
\boldsymbol{x}^\mathrm{T}[\boldsymbol{x}-\boldsymbol{P}(\boldsymbol{x})]
\end{aligned}
\end{equation}
here, $\boldsymbol{C}=[\boldsymbol{U}_{\boldsymbol{\Theta}}(\boldsymbol{z})+\boldsymbol{u}]$. Eq. (\ref{g5-00001}) is a classic RED objective, representing a denoising of the target $\boldsymbol{C}$.

Here, we assume that $\boldsymbol{P}(\boldsymbol{x})$ of DRUNet satisfies \textit{RED conditions} \footnote{In the following Section \ref{4}, we give detailed validation about how our empolyed deep denoiser DRUNet meets these three \textit{RED conditions}, and how to derive the differential form Eq. (\ref{red_add1}) of the RED term by satisfying the \textit{RED conditions.}}. When the gradient expression of $\rho\boldsymbol{(x)}$ satisfies Eq. (\ref{red_add1}), the above Eq. (\ref{g5-00001}) can be efficiently solved through fixed-point method by zeroing the derivative of Eq. (\ref{g5-00001}). Based on the above, by setting the gradient of Eq. (\ref{g5-00001}) to zero, we have:
\begin{equation}\label{g5-1}
\begin{aligned}
    0=\mu(\boldsymbol{x}-\boldsymbol{C})+\lambda[\boldsymbol{x}-\boldsymbol{P}(\boldsymbol{x})]
\end{aligned}
\end{equation}
Eq. (\ref{g5-1}) can be solved iteratively by fixed-point method:
\begin{equation}\label{g6}
  \boldsymbol{x}^{k+1}=\frac{1}{\lambda+\mu}[\lambda \boldsymbol{P}(\boldsymbol{x}^{k})+\mu \boldsymbol{C}]
\end{equation}
$k$ is the iteration number.

\subsection{Update of sub-problem}

As last, Lagrangian multiplier $\boldsymbol{u}$ is updated as:
\begin{equation}\label{G9}
  \boldsymbol{u}^{k+1}=\boldsymbol{u}^{k}-\boldsymbol{x}+\boldsymbol {U}_{\boldsymbol{\theta}}(\boldsymbol{z})
\end{equation}

\noindent \textbf{Algorithm 1} summarizes the pseudo-code of Net2Net.

\begin{algorithm}
    \caption{Our proposed Net2Net for real noise removal}\label{algo:2} 
        \begin{algorithmic} 

 \REQUIRE $\boldsymbol{y}$, $\lambda$ and $\mu$.     
\STATE \textbf{Initialization}: Set $\boldsymbol{u}_{0}=\boldsymbol{0}$, $\boldsymbol{x}_{0}=\boldsymbol{y}$, noise vector $\boldsymbol{z}$. 
   \FOR{$k=0,1,..., Iter$}  
    \STATE Update $\boldsymbol{\Theta}_{}^{k+1}$ by solving Eq. (\ref{g4}). 
    \STATE Update $\boldsymbol{x}_{}^{k+1}$ by solving Eq. (\ref{g6}). 
    \STATE Update $\boldsymbol{u}_{}^{k+1}$ by solving Eq. (\ref{G9}). 
   \ENDFOR
           \RETURN The recovered image $\hat{\boldsymbol{x}}$.
      \end{algorithmic}
\end{algorithm}

\section{Meet RED Conditions}\label{4}
In Section \ref{4.3.2}, for updating $\boldsymbol{x}$ sub-problem, we assume that denoising engine $\boldsymbol{P}(\boldsymbol{x})$ of DRUNet satisfies \textit{RED conditions}, so that the gradient expression Eq. (\ref{red_add1}) is established. Fixed-point method is employed in solving Eq. (\ref{g6}). In this Section, we verify that DRUNet satisfies the three required \textit{RED conditions}. And how these three conditions lead to the gradient expression Eq. (\ref{red_add1}) holds.

\subsection{RED Condition 1: Differentiability}

\begin{table*}[h]
\centering
\setlength{\tabcolsep}{4pt}
\renewcommand{\arraystretch}{1.2}
\caption{ The
detailed architecture of DRUNet.}
\label{table_drunet}
\begin{tabular}{llll}
\toprule
\textbf{Name} & & \textbf{Type}   & \textbf{Output} \\ \midrule

Input &&& 1 $\times$ 4 $\times$ 512 $\times$ 512  \\\hline

Conv &&Conv2d (3,3)& 1 $\times$ 64 $\times$ 512 $\times$ 512 \\\hline
4 Residual Blocks  &\multirow{6}{*}{{\textbf{Downscaling}}}&Conv2d (3,3)-ReLU-Conv2d (3,3) & 1 $\times$ 64 $\times$ 512 $\times$ 512\\

SConv   &&Strided convolution downscaling (2,2) & 1 $\times$ 128 $\times$ 256 $\times$ 256 \\

4 Residual Blocks &&Conv2d (3,3)-ReLU-Conv2d (3,3)& 1 $\times$ 128 $\times$ 256 $\times$ 256\\
SConv   &&Sonv2d (2,2)& 1 $\times$ 256 $\times$ 128 $\times$ 128  \\

4 Residual Blocks &&Conv2d (3,3)-ReLU-Conv2d (3,3)& 1 $\times$ 256 $\times$ 128 $\times$ 128\\
SConv   &&SConv2d (2,2)&1 $\times$ 512 $\times$ 64 $\times$ 64  \\\hline

4 Residual Blocks &&Conv2d (3,3)-ReLU-Conv2d (3,3)&1 $\times$ 512 $\times$ 64 $\times$ 64 \\\hline

TConv  &\multirow{6}{*}{{\textbf{Upscaling}}}&TConv2d (2,2) & 1 $\times$ 256 $\times$ 128 $\times$ 128 \\
4 Residual Blocks &&Conv2d (3,3)-ReLU-Conv2d (3,3)&1 $\times$ 256 $\times$ 128 $\times$ 128 \\

TConv  &&TConv2d (2,2)& 1 $\times$ 128 $\times$ 256 $\times$ 256  \\
4 Residual Blocks &&Conv2d (3,3)-ReLU-Conv2d (3,3)&1 $\times$ 128 $\times$ 256 $\times$ 256  \\

TConv  &&TConv2d (2,2)& 1 $\times$ 64 $\times$ 512 $\times$ 512   \\
4 Residual Blocks &&Conv2d (3,3)-ReLU-Conv2d (3,3)& 1 $\times$ 64 $\times$ 512 $\times$ 512 \\\hline

Conv &&Conv2d (3,3)& 1 $\times$ 3 $\times$ 512 $\times$ 512  \\\bottomrule

\end{tabular}
\end{table*}

DRUNet is a UNet \cite{ronneberger2015u} architecture, with residual connection \cite{he2016deep}. The detailed architecture of DRUNet is listed in Table \ref{table_drunet}. Spatial downsampling is performed using 2$\times$2 convolutions with stride 2, while spatial upsampling leverage 2$\times$2 transposed convolutions with stride 2. The number of channels in each layer are 64, 128, 256 and 512, respectively. 4 successive residual blocks are adopted in the downscaling and upscaling of each scale. Each residual block only contains one ReLU activation function. Importantly, DRUNet is bias-free (BF), which means no bias is used in all the Conv, SConv and TConv layers. 

Common differentiable operations include convolution (i.e., Conv, SConv and TConv) and ReLU. The residual connections also maintain differentiability as they only involve element-wise addition, a differentiable operation. Regarding their respective differentiability, when these operations are used sequentially within DRUNet, the linear combination $\boldsymbol{P}(\boldsymbol{x})$ is differentiable.

\subsection{RED Condition 2: Homogeneity}

For a given noisy input image $\boldsymbol{x}$, the function of DRUNet $\boldsymbol{P}(\cdot)$ computed by a general denoising neural network with $L$ layers can be expressed as: 
\begin{equation}\label{g4-1}
\begin{aligned}
    \boldsymbol{P}(\boldsymbol{x})=\boldsymbol{W}_L\boldsymbol{R}(\boldsymbol{W}_{L-1}\cdots\boldsymbol{R}(\boldsymbol{W}_1 \boldsymbol{x}+\boldsymbol{b}_1)+\cdots\boldsymbol{b}_{L-1})+\boldsymbol{b}_{L}
\end{aligned}
\end{equation}
where $\boldsymbol{W}_i$ indicates the convolutional layer, $\boldsymbol{b}_i$ is the additive constants of net bias, with $i \in [1, L]$. $\boldsymbol{R}$ represents the fixed activation pattern ReLU.

As mentioned before, DRUNet is bias-free, and the net action is strictly linear. Eq. (\ref{g4-1}) can be further reformulated as:
\begin{equation}\label{g4-2}
\begin{aligned}
    \boldsymbol{P}(\boldsymbol{x})=\boldsymbol{W}_{L} \boldsymbol{R}(\boldsymbol{W}_{\boldsymbol{L}-1}\cdots\boldsymbol{R}(\boldsymbol{W}_1 \boldsymbol{x}))
\end{aligned}
\end{equation}

\textit{Scale-Equivariance Property \cite{mohan2019robust}}:
Let $\boldsymbol{P}_{BF}(\cdot)$ be a feedforward neural network with ReLU activation functions and no additive constant terms in any layer. For any input $\boldsymbol{x}$ and any non-negative constant $\alpha$:
\begin{equation}\label{g4-3}
\begin{aligned}
    \boldsymbol{P}_{BF}(\alpha \boldsymbol{x})=\alpha \boldsymbol{P}_{BF}(\boldsymbol{x})
\end{aligned}
\end{equation}

Proof: Rectifying operator ReLU, termed as $\boldsymbol{R}$, sets to zero any negative entries in its input. Multiplying by a non-negative constant does not change the sign of the entries of a vector, so for any $\boldsymbol{x}$ with the right dimension and any $\alpha \textgreater 0$, we have $\boldsymbol{R}(\alpha \boldsymbol{x})=\alpha \boldsymbol{R}(\boldsymbol{x})$, thus:
\begin{equation}\label{g4-4}
\begin{aligned}
    \boldsymbol{P}_{BF}(\alpha \boldsymbol{x})&=\boldsymbol{W}_{L} \boldsymbol{R}(\boldsymbol{W}_{L-1}...\boldsymbol{R}(\boldsymbol{W}_1 \alpha \boldsymbol{x}))\\&=\alpha \boldsymbol{W}_{L} \boldsymbol{R}(\boldsymbol{W}_{L-1}...\boldsymbol{R}(\boldsymbol{W}_1 \boldsymbol{x}))\\&=\alpha \boldsymbol{P}_{BF}(\boldsymbol{x})
\end{aligned}
\end{equation}

\noindent Note that Eq. (\ref{g4-4}) also holds for $\boldsymbol{P}(\cdot)$ with skip or residual connections where the feature maps are concatenated or added, because both of these operations are linear. From the above, we can see that the pre-trained deep prior DRUNet is an unbiased architecture, activated by ReLU and using residual and skip connection structures. So for $\boldsymbol{P}(\cdot)$ of DRUNet, \textit{Lemma 5.1} has proven to hold for $\alpha \textgreater 0$, it follows that the \textit{Lemma 5.1}
is also valid for $\alpha \textgreater 1$. So \textit{RED Condition 2} is hold.

\begin{table*}[h]
\setlength{\tabcolsep}{2pt}
\renewcommand{\arraystretch}{1}
\caption{SSIM \cite{1284395} and MSE between $\boldsymbol{P}[(1+\epsilon)\boldsymbol{x}]$ and $(1+\epsilon)\boldsymbol{P}(\boldsymbol{x})$ with $\epsilon=0.01$ on Set12 \cite{8365806}, where $\boldsymbol{x}$ indicates image corrupted by AWGN with noise level $\sigma_{n}=5$.}
\label{table_home}
\centering
\begin{tabular}{lccccccccccccc}
  \hline
Image   &Man & Hous. & Pepp. & Star. & Mona.& Airp.& Parr.& Lena & Barb. & Boat & Man& Coup.&Avg \\ \hline 
SSIM &0.9982&0.9986 &0.9989&0.9991&0.9991&0.9992&0.9984&0.9981&0.9987&0.9983&0.9985&0.9987&0.9986\\\hline
MSE &0.4e-5&0.3e-5 &0.4e-5&0.4e-5&0.4e-5&0.2e-5&0.4e-5&0.4e-5&0.4e-5&0.5e-5&0.5e-5&0.4e-5&0.4e-5\\\hline
\end{tabular}
\end{table*}

Moreover, following the same setting as \cite{doi:10.1137/16M1102884}, we also provide experimental evidence to validate \textit{RED Condition 2}. Testing image \textit{pepper} is corrupted by AWGN with level $\sigma_{n}=50$ to be $\boldsymbol{x}$. $\epsilon$ is set as 0.001. Table \ref{table_home} shows the evaluation of homogeneity, SSIM \cite{1284395} and MSE are used to measure the similarity between $\boldsymbol{P}[(1+\epsilon)\boldsymbol{x}]$ and $(1+\epsilon)\boldsymbol{P}(\boldsymbol{x})$, the average SSIM and MSE between these two terms are 0.9986 and 0.4e-5, very close to 1 and 0, respectively. The above verify the homogeneity property of $D(\boldsymbol{x})$ from experimental perspective.

\subsection{RED Condition 3: Jacobian Symmetry}

As the deep model $\boldsymbol{P}(\cdot)$ of DRUNet employed in our framework does not have a general
closed-form formulation. It is difficult to directly prove the symmetry of Jacobian. Following prior work \cite{Reehorst2018RegularizationBD,yuan2024mixed,faye2024regularization}, normalized error metric (NEM) is used to verify the symmetry property of Jacobian matrix. NEM is defined as:
\begin{equation}\label{g-7-19-5}
\begin{aligned}
    e(\boldsymbol{x})=\frac{\| J \boldsymbol{P}(\boldsymbol{x})-[J \boldsymbol{P}(\boldsymbol{x})_{}]^{T} \|_{F}^{2}}{\| J \boldsymbol{P}(\boldsymbol{x}) \|_{F}^{2}}
\end{aligned}
\end{equation}
here, $[J\boldsymbol{P}(\boldsymbol{x})]_{i,n} = [\boldsymbol{P}_{i}({\boldsymbol{x}+\varrho \boldsymbol{e}_{n}})-\boldsymbol{P}_{i}({\boldsymbol{x}-\varrho \boldsymbol{e}_{n}})]/{2\varrho}$. $\varrho$ is small constant and $\boldsymbol{e}_{n}$ is the $n$-th column of identity matrix. The smaller NEM value, the closer of $\boldsymbol{P}(\boldsymbol{x})$ is to symmetry.

\begin{table}[!ht]\tiny
\setlength{\tabcolsep}{8pt}
\renewcommand{\arraystretch}{1}
\caption{Average NEM performance of common denoisers on Set12 \cite{8365806}. The smaller NEM value, the closer of $\boldsymbol{P}(\boldsymbol{x})$ is to symmetry.}
\small
\label{table_jacobian}
\centering
\begin{tabular}{lccccc}
  \hline
$\boldsymbol{P}(\boldsymbol{x})$&NLM&BM3D  &TNRD &DnCNN &DRUNet  \\ \hline 
$e(\boldsymbol{x})$ &0.7638&0.7039&0.0259&0.0116&0.0004  \\\hline
\end{tabular}
\end{table}

Table \ref{table_jacobian} lists the average NEM of several denoisers: NLM \cite{1467423}, BM3D \cite{4378954}, TNRD \cite{7527621}, DnCNN \cite{7839189}, FFDNet \cite{8365806} and DRUNet \cite{945431111} under $\sigma_{n}=50$ and $\varrho=0.001$.  It can be observed that $e(\boldsymbol{x})$ of DRUNet is the smallest, closest to 0. This verifies the symmetry of the Jacobian matrix of DRUNet. Besides, denoisers, such as NLM, BM3D, TNRD, and DnCNN are characterized by non-symmetric Jacobian. Yet, RED-based restoration methods \cite{doi:10.1137/16M1102884,Reehorst2018RegularizationBD,faye2024regularization} are shown to empirically converge and to reach excellent performance when solving various inverse problems even these RED conditions are partially satisfied.

\subsection{How RED Conditions Lead to RED Gradient Holds}
Under \textit{RED conditions}, RED term Eq. (\ref{red}) is differentiable and convex, and the gradient  $\nabla\rho\boldsymbol{(x)}=\boldsymbol{x}-\boldsymbol{P}(\boldsymbol{x})$. Based on it, first-order optimization method fixed-point iteration is employed for updating $\textbf{x}$ sub-problem (see Section \ref{4.3.2}).
Now, we present a proof demonstrating how RED Gradient is hold under the condition that $\boldsymbol{P}(\boldsymbol{x})$ of DRUNet satisfies above three \textit{RED conditions}.

First, we present two lemmas along with corresponding proofs.

\textit{RED general gradient \cite{Reehorst2018RegularizationBD}}
For RED term defined in Eq. (\ref{red}), its gradient can be expressed as the following general form:
\begin{equation}\label{g4-3}
\begin{aligned}
    \nabla \rho\boldsymbol{(x)}
&= \boldsymbol{x} - \frac{1}{2} P(
\boldsymbol{x}) 
   - \frac{1}{2} [J P(\boldsymbol{x})]^T \boldsymbol{x}
\end{aligned}
\end{equation}
\label{eq:gradRED}

Proof: We first present the following definition, we denote the $i$th component of $\boldsymbol{P}(\boldsymbol{x})$ by $P_{i}(\boldsymbol{x})$ with $i \in [1,N]$, thus the gradient of $P_{i}(\boldsymbol{\cdot})$ at $\boldsymbol{x}$ is expressed as:
\begin{equation}\label{g-31}
\begin{aligned}
\nabla P_{i}(\boldsymbol{x})=\bigg [\frac{\partial P_{i}(\boldsymbol{x})}{\partial x_1}, \frac{\partial P_{i}(\boldsymbol{x})}{\partial x_2} ,\cdots, \frac{\partial P_{i}(\boldsymbol{x})}{\partial x_N}\bigg]
\end{aligned}
\end{equation}
and the Jacobian matrix of $\boldsymbol{P}(\boldsymbol{\cdot})$ at $\boldsymbol{x}$ can be expressed as: 
\begin{equation}\label{g-31}
\begin{aligned}
J \boldsymbol{P}(\boldsymbol{x})=\bigg [\nabla P_{1}(\boldsymbol{x}), \nabla P_{2}(\boldsymbol{x}), \cdots, \nabla P_{N}(\boldsymbol{x}) \bigg]^T=\nabla_{\boldsymbol{x}}P(\boldsymbol{x})
\end{aligned}
\end{equation}
thus, the gradient of RED term can be obtained by:
\begin{align}
\lefteqn{
\frac{\partial \rho\boldsymbol{(x)}}{\partial x_n}
= \frac{\partial}{\partial x_n} \frac{1}{2}
\sum_{i=1}^N \big( x_i^2 - x_i P_i(\boldsymbol{x}) \big) }\\
&= \frac{1}{2} \frac{\partial}{\partial x_n} \left(
x_n^2 
- x_n P_n(\boldsymbol{x}) 
+ \sum_{i\neq n} x_i^2 
- \sum_{i\neq n} x_i P_i(\boldsymbol{x}) \right)\\
&= \frac{1}{2} \left( 
2x_n 
- D_n(\boldsymbol{x}) 
- x_n 
\frac{\partial P_n(\boldsymbol{x})}{\partial x_n} 
- \sum_{i\neq n} x_i 
\frac{\partial P_i(\boldsymbol{x})}{\partial x_n} 
\right) \\
&= x_n - \frac{1}{2} P_n(\boldsymbol{x}) - \frac{1}{2} \sum_{i=1}^N x_i 
\frac{\partial P_i(\boldsymbol{x})}{\partial x_n}  
\label{eq:partialRED} \\
&= x_n - \frac{1}{2} P_n(\boldsymbol{x}) 
- \frac{1}{2} \left[ [J\boldsymbol{P}(\boldsymbol{x})]^T \boldsymbol{x} \right]_n
\end{align}
using the definition of $J f(\boldsymbol{x})$ defined from Eq. (\ref{g-31}).
Collecting $\{\frac{\partial \rho\boldsymbol{(x)}}{\partial x_n}\}_{n=1}^N$ into the gradient vector yields Eq. (\ref{g4-3}). The proof completes.

Local homogeneity \cite{doi:10.1137/16M1102884}
Suppose that pre-trained denoiser $\boldsymbol{P}(\cdot)$ is locally homogeneous. 
Then $[J \boldsymbol{P}(\boldsymbol{x})]\boldsymbol{x} =\boldsymbol{P}(\boldsymbol{x})$, where $J \boldsymbol{P}(\cdot)$ is the Jacobian matrix of $\boldsymbol{P}(\cdot)$.

proof: The proof is based on differentiability and avoids the need to define a directional derivative.
From \cite{Reehorst2018RegularizationBD,rudin1964principles}, we have 
\begin{align}
0 
&= \lim_{\epsilon \rightarrow 0} \frac{\|\boldsymbol{P}(\boldsymbol{x}+\epsilon \boldsymbol{x})-\boldsymbol{P}(\boldsymbol{x})- [J \boldsymbol{P}(\boldsymbol{x})]\epsilon \boldsymbol{x}\|}{\|\epsilon\boldsymbol{x}\|} \\
&= \lim_{\epsilon \rightarrow 0} \frac{\|(1+\epsilon)\boldsymbol{P}(\boldsymbol{x})-\boldsymbol{P}(\boldsymbol{x}) - [J \boldsymbol{P}(\boldsymbol{x})]\epsilon \boldsymbol{x}\|}{\|\epsilon\boldsymbol{x}\|} 
\label{eq:LH2}\\
&= \lim_{\epsilon \rightarrow 0} \frac{\|\boldsymbol{P}(\boldsymbol{x}) - [J \boldsymbol{P}(\boldsymbol{x})]\boldsymbol{x}\|}{\|\boldsymbol{x}\|} 
\label{eq:LH3}
\end{align}
where $\epsilon$ is a small constant. Eq.(\ref{eq:LH2}) follows from the proven RED \textit{Condition 2: Homogeneity}. Eq. (\ref{eq:LH3}) implies that $[J \boldsymbol{P}(\boldsymbol{x})]\boldsymbol{x} =\boldsymbol{P}(\boldsymbol{x})$.

Based on the \textit{Lemma 5.3} and \textit{
RED Condition 3: Jacobian Symmetry}, the general form of RED Eq. (\ref{g4-3}) (namely \textit{Lemma 5.2}, Eq. (\ref{eq:mass-1})) can be repressed as: 
\begin{align}
\nabla \rho\boldsymbol{(x)} 
&= \boldsymbol{x} - \frac{1}{2} \boldsymbol{P}(
\boldsymbol{x}) 
   - \frac{1}{2} [J \boldsymbol{P}(\boldsymbol{x})]^T \boldsymbol{x}\label{eq:mass-1}
\\&=\boldsymbol{x} - \frac{1}{2} \boldsymbol{P}(
\boldsymbol{x}) 
   - \frac{1}{2} [J \boldsymbol{P}(\boldsymbol{x})] \boldsymbol{x}\label{eq:mass-2}
\\&=\boldsymbol{x} - \frac{1}{2} \boldsymbol{P}(
\boldsymbol{x}) 
   - \frac{1}{2} \boldsymbol{P}(
\boldsymbol{x})\label{eq:mass-3}
\\&=\boldsymbol{x} - \boldsymbol{P}(
\boldsymbol{x})\label{eq:mass-4}
\end{align}
where Eq. (\ref{eq:mass-2}) follows from \textit{
RED Condition 3: Jacobian Symmetry}, Eq. (\ref{eq:mass-3}) follows from \textit{Lemma 5.3}.

We have provided a proof of the equivalence between \textit{Lemma 5.2} and Eq. (\ref{red_add1}). Thus, when the former holds, it follows that the latter also holds. Till now, we provide the detailed proof that how RED conditions lead to RED gradient holds. Based on the above, our proposed method are guaranteed to converge to a global optimum solution.  


\section{Experiment}\label{5}

In this part, we elaborate on the used datasets, baselines, the evaluation metrics, and the implementation details.

\subsection{Performance on Natural Images}

\subsubsection{Implementation Details}

\textbf{Compared Methods.} For comprehensive comparisons, we compare the proposed Net2Net with 18 methods which are divided into 3 groups. To be specific, 6 Full-Supervised Methods: DnCNN \cite{7839189}, FFDNet \cite{8365806}, CBDNet \cite{8954448}, RIDNet \cite{9010324}, AINDNet \cite{9157160}, InvDN \cite{9577357}.
6 Self-Supervised Methods: S2S \cite{9157420}, R2R \cite{9577798}, CVF-SID \cite{Neshatavar_2022_CVPR}, AP-BSN \cite{9878719}, ZS-N2N \cite{10203304}, SDAP \cite{Pan_2023_ICCV}.
6 Un-Supervised Methods: BM3D \cite{4378954}, WNNM \cite{6909762}, DIP \cite{Ulyanov2017DeepIP}, SB \cite{54}, NN \cite{zheng2021unsupervised}, PD \cite{zhou2020awgn}, ScoreDVI \cite{cheng2023score}. For
the compared methods without the corresponding results, we carry out them by using the source codes provided by the authors and adopting their parameter settings.

\textbf{Compared Metrics.} To evaluate the restoration quality, we use the PSNR and SSIM \cite{1284395} metrics under the RGB color space.  
PSNR measures the quality of a reconstructed image by calculating the ratio between the maximum possible signal power and the power of corrupting noise. SSIM compares the structural similarity between the reconstructed and the original image.

\textbf{Dataset.} We conduct experiment on 3 real datasets: CC \cite{7780555}, PolyU \cite{xu1804real}, and RN15 \cite{lebrun2015noise}. To be specific, \textbf{1. CC} includes noisy images of scenes captured by Canon 5D Mark 3, Nikon D600 and Nikon D800 cameras. The real noisy images are collected under controlled indoor environment. Each scene is shot 500 times under the same camera and setting. The mean image of the 500 shots is taken as the truth. 15 images of size 3 $\times$ 512 $\times$ 512 are cropped to evaluate various methods. \textbf{2. PolyU} is a more comprehensive dataset taken from 40 versatile scenes in different lighting conditions. To include more camera settings, each image is captured with 6 difference ISO factors. Each scene is captured 500-1000 times and the ones with spatial misalignment and luminance discrepancy are removed. Multiple samples of the same scene are averaged and taken as the clean ground truth. \textbf{3. RN15} comprises 15 real noisy images. The images in RNI15 cover a variety of noise types including low-light images from smartphones, old photographs, aerial images, etc. Due to the absence of clean ground truths, RNI15 is only used for qualitative evaluation purposes.

\subsubsection{Experimental Results}

\begin{figure*}[!h]
\centerline{\includegraphics[width=1\linewidth]{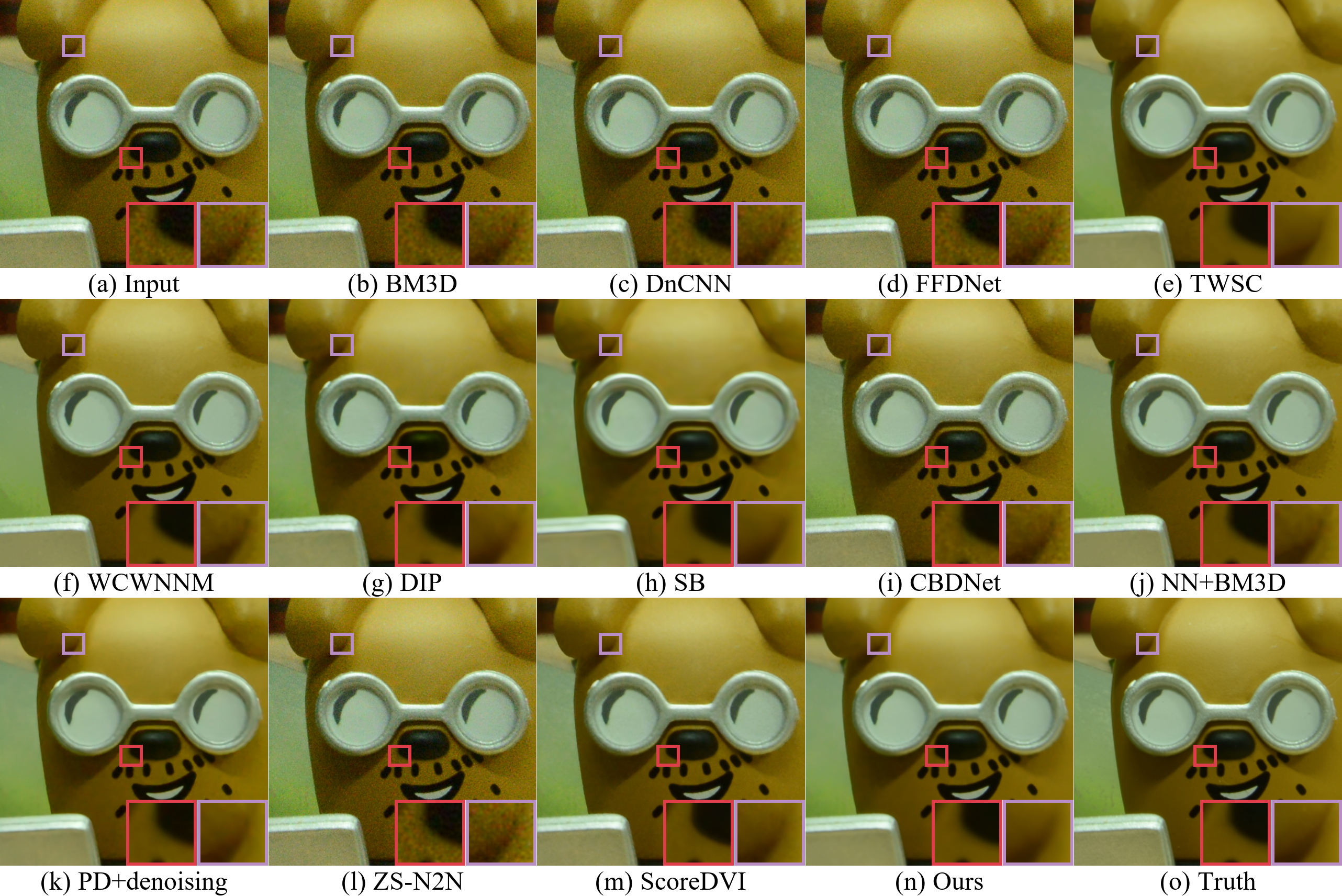}}
 \caption{Visual comparisons between our method and other state-of-the-art denoising methods on the dataset. Our method eliminates complex noise effectively while retaining more structural content and texture, leading to artifact-free result.} 
\label{cc_13}
\end{figure*}
\begin{figure*}[!h]
\centerline{\includegraphics[width=1\linewidth]{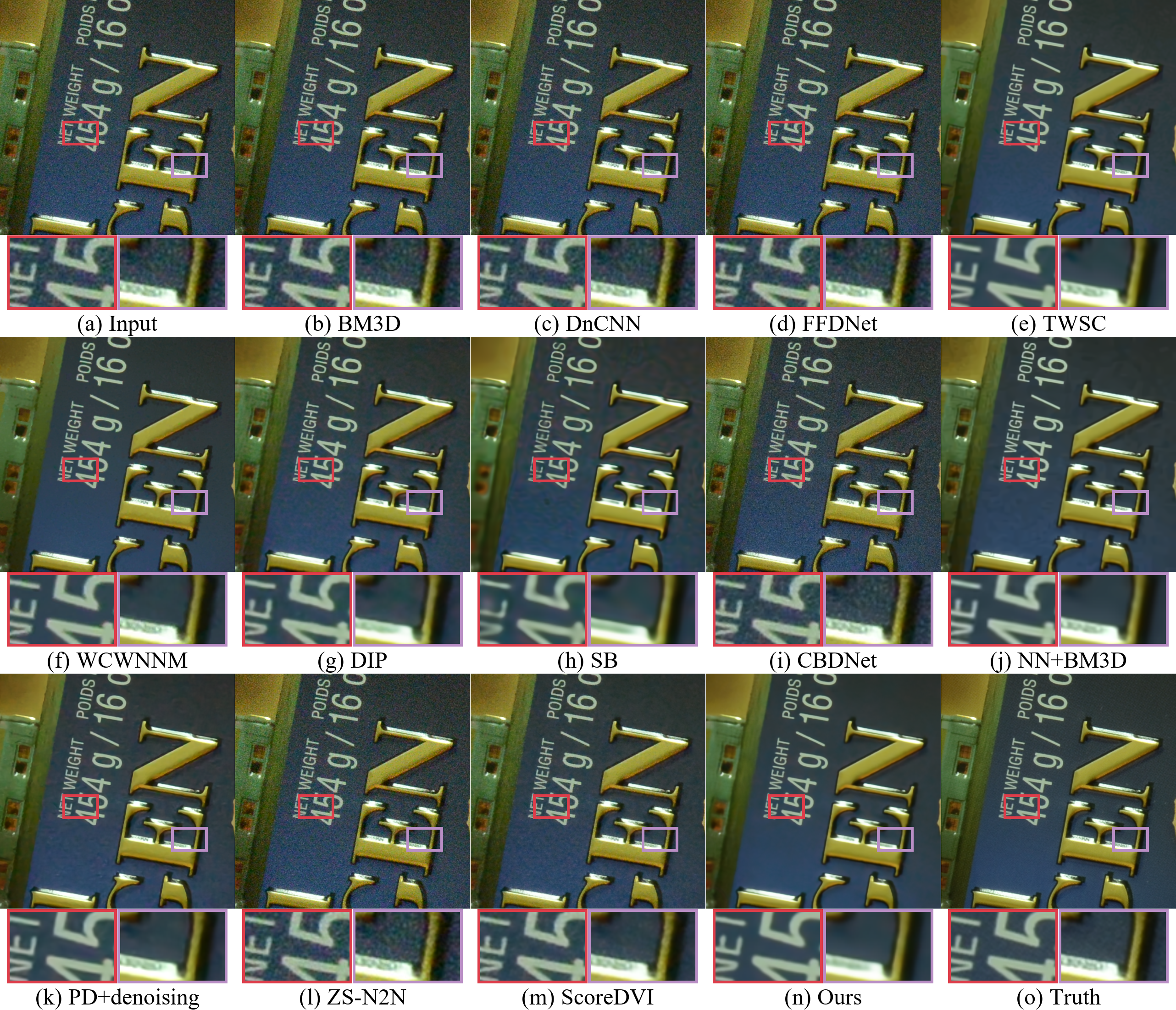}}
 \caption{Visual comparisons between our method and other state-of-the-art denoising methods on the dataset. Our method eliminates complex noise effectively while retaining more structural content and texture, leading to artifact-free result.} 
\label{cc_10}
\end{figure*}

Figs. \ref{cc_13} and \ref{cc_10} shows a visual comparison of the denoising results of different algorithms. Net2Net performs excellently on real noisy images, effectively removing complex noise while also preserving detailed image structure. In contrast, self-supervised algorithms such as ScoreDVI and ZS-N2N exhibit significant residual noise, resulting in artifacts such as blurred edges, which compromise the perceived naturalness of the image. While WCWNNM and TWSC demonstrate some denoising capabilities, they are prone to oversmoothing when processing complex textured areas, leading to loss of image detail. Furthermore, the supervised algorithms DnCNN, FFDNet, and CBDNet exhibit unsatisfactory denoising results due to domain shift between training and test data. In contrast, Net2Net, leveraging the collaborative optimization mechanism between an untrained deep structural prior and a pretrained deep denoising prior (DRUNet), effectively removes noise while better restoring texture details and edge structure in degraded images. Based on the above analysis, Net2Net not only achieves excellent performance in objective indicators such as PSNR and SSIM in real noise removal, but also presents a more natural and clear visual effect, verifying Net2Net's ability to avoid overfitting of noise in actual degradation scenarios.


Table \ref{table_cc15} and Table \ref{table_3} show the qualitative results of different denoising algorithms on the CC and PolyU datasets, demonstrating that the algorithms presented in this chapter demonstrate competitive performance. Specifically, compared to denoising algorithms based on physical prior models, Net2Net achieves improvements in average PSNR and SSIM over BM3D, WNNM, WCWNNM, and TWSC by (6.47\%, 6.83\%, 1.42\%, 4.57\%) and (6.51\%, 2.88\%, 0.42\%, 0.42\%), respectively. This indicates that the physical prior model does not accurately describe the distribution of real noise, resulting in limited effectiveness in removing real noise. Compared with supervised learning-based denoising algorithms, Net2Net improves the average PSNR and SSIM by (6.27\%, 10.52\%, 3.23\%, 5.67\%, 2.66\%, 4.32\%, 6.30\%, 6.21\%, 4.46\%) over IRCNN, DnCNN, FFDNet, CBDNet, RIDNet, ADNetB, AINDNet, MWDCNN, and Restormer, respectively. Supervised learning denoising algorithms, which rely heavily on training data, experience performance degradation and poor generalization when faced with domain shift between training and test data. Net2Net, on the other hand, demonstrates superior denoising capabilities and generalization by jointly optimizing two types of network priors (pre-trained deep denoising priors and network structure priors) within the RED framework. Compared to self-supervised denoising algorithms, Net2Net improves average PSNR and SSIM by 3.93\%, 5.57\%, 11.03\%, 13.81\%, and 3.2\%, respectively, over R2R, B2U, ZS-N2N, SDAP, and ScoreDVI. These algorithms rely on prior assumptions about noisy images, but in real-world scenarios, such as R2R, B2U, and ZS-N2N, the zero-mean noise assumption fails, hindering their superior denoising performance. In contrast, Net2Net demonstrates greater robustness and generalization in realistic, degraded noise scenarios, demonstrating superior qualitative results.

\begin{table*}[!h]
\setlength{\tabcolsep}{1.1pt}
\renewcommand{\arraystretch}{1.1}
\caption{PSNR scores of different methods on CC(Nam) \cite{7780555}.}
\label{table_cc15}
\centering
\begin{tabular}{llccccccccccccccccc}
\toprule
\textbf{Method}&Venue&Img1&Img2&Img3&Img4&Img5&Img6&Img7&Img8&Img9&Img10&Img11&Img12&Img13&Img14&Img15&\textbf{Avg.}\\\midrule

{DnCNN}\cite{7839189}&\tiny{\textit{TIP'2017}}&37.26&34.13&34.09&33.62&34.48&35.41&37.95&36.08&35.48&34.08&33.70&33.31&29.83&30.55&30.09&33.86\\

{FFDNet}\cite{8365806}&\tiny{\textit{TIP'2018}}&39.40&37.02&36.53&34.97&36.73&41.02&38.66&41.53&38.80&40.15&37.61&41.18&34.13&33.66&33.69&37.68 \\


{CBDNet}\cite{8954448} & \tiny{\textit{CVPR'2019}}&38.62  &35.59&35.29&35.01&36.13&39.14&38.65&40.13&38.34&37.60&36.42&37.26&31.29&32.20&31.63&36.22\\

{RIDNet}\cite{9010324} &\tiny{\textit{ICCV'2019}} &37.63  &{37.28}&{37.75}&{34.55}&35.99&38.62&39.22&39.67&39.04&38.28&37.18&38.85&32.75&33.24&32.89&36.73 \\

{AINDNet}\cite{9157160}&\tiny{\textit{CVPR'2020}}& 37.03  &37.19&{37.24}&35.45&36.79&38.65&39.61&38.84&38.68&37.50&37.58&38.35&33.81&33.68&32.87&36.88\\

{ADNet}\cite{tian2020attention} &\tiny{\textit{NN'2020}}&35.96&36.11&34.49&33.94&34.33&38.87&37.61&38.24&36.89&37.20&35.67&38.09&32.24&32.59&33.14&35.69\\

{MWDCNN}\cite{tian2023multi} &\tiny{\textit{PR'2023}}&36.97&36.01&34.80&33.91&34.88&37.02&37.93&37.49&38.44&37.10&36.72&37.25&32.24&32.56&32.76&35.74 \\


{Restormer}\cite{zamir2022restormer}&\tiny{\textit{CVPR'2022}} & 36.88 & 36.71& 35.65&34.41 &36.10 &38.09&38.70&38.93&39.31 &37.83&37.82&37.40 &31.94&32.70&32.15&36.31 \\\hline 

R2R\cite{9577798} &\tiny{\textit{CVPR'2021}}&
39.11&35.45&35.71&34.62&36.37&37.44&37.97&37.66&36.26&37.87&34.19&39.90&32.65&31.09&32.27&35.90   \\


B2U\cite{Wang_2022_CVPR}&\tiny{\textit{CVPR'2022}}&37.81&35.14&34.63&33.61&34.72&36.22&37.95&36.64&36.18&36.78&34.93&35.62&32.23&30.90&31.22& 34.97  \\

{ZS-N2N}\cite{10203304} &\tiny{\textit{CVPR'2023}}&37.30&  34.21&34.13&33.33&33.84&35.07&35.57&35.89&34.99&33.39&32.94&33.18&29.80&30.08&29.97&33.58          \\

{SDAP}\cite{Pan_2023_ICCV}&\tiny{\textit{ICCV'2023}}&35.28&33.02&35.63&34.13&31.98&31.99&38.13&38.69&31.12&32.38&36.56&32.76&32.51&27.41&32.59&33.61    \\\hline

{BM3D}\cite{4378954}&\tiny{\textit{TIP'2007}} &  39.76&36.40&36.37&34.18&35.07&37.13&36.81&37.76&37.51&35.05&34.07&34.42&31.13&31.22&30.97&35.19 \\ 

WNNM\cite{6909762}  &\tiny{\textit{CVPR'2014}}&37.51&33.86&31.43&33.46& 36.09&39.86&36.35&39.99&37.15&38.60&36.04&39.73&33.29&31.16&31.98&35.77 \\

{MCWNNM}\cite{8237387}&\tiny{\textit{ICCV'2017}} &   {41.20}& {37.25}& {37.06}&{35.54} &{37.03} &{39.56}&{39.26}&{41.45}&{39.54} &{38.94}&{37.40}& {39.42}&{34.85}&{33.97}&{33.97}&{37.76} \\ 

{TWSC}\cite{xu2018trilateral}&\tiny{\textit{ECCV'2018}}&40.55&35.92&35.15&35.36&37.09&41.13&39.36&41.91&38.81&40.27&37.22&42.09&35.53&34.15&33.93&37.90\\


{DIP}\cite{Ulyanov2017DeepIP} &\tiny{\textit{IJCV'2020}}&38.49&35.82&34.29&33.64&36.19&41.60&37.00&41.39&37.83&39.63&36.85&43.41&35.07&33.05&33.76&37.20    \\

{SB}\cite{54}&\tiny{\textit{IJCV'2022}}&39.67&35.75&36.42&34.77&36.27&42.05&37.63&40.45&36.86&38.94& 36.55&42.75&33.42&32.84&32.66&37.14         \\

DRP-DIP\cite{Li_2023_CVPR}& \tiny{\textit{CVPR'2023}} & 38.68&34.53&35.37&34.23&35.41&39.55&38.05&38.91&37.16&39.26&35.29&37.89&32.80&32.38&32.50&36.13\\

NN+BM3D\cite{zheng2021unsupervised}& \tiny{\textit{ICLR'2021}}&40.58&37.17&36.87&35.29&36.91&41.50&39.12&41.89&39.49&40.09&37.37&43.09&34.76&34.18&34.24&38.17 \\

PD\cite{zhou2020awgn}&\tiny{\textit{AAAI'2020}}&37.53 &34.72&34.22&33.61&34.49&35.57&36.30&36.37&36.25&34.83&33.98&33.49&29.82&30.57&30.09&34.12    \\


ScoreDVI\cite{cheng2023score} &\tiny{\textit{ICCV'2023}}
&39.35
&36.32
&36.28
&35.28
&36.62
&40.38
&38.49
&40.78
&38.26
&39.05
&36.63
&39.44
&33.88
&32.97
&33.10
&37.12            \\\hline

{Net2Net} & &\textbf{41.25}&37.11&37.16&35.36&37.02&42.09&39.10&42.29&39.28&40.23&37.47&45.21&35.39&33.70&34.05&38.45\\\bottomrule

\end{tabular}
\end{table*}

\begin{table*}[!h]
\setlength{\tabcolsep}{1.2pt}
\renewcommand{\arraystretch}{1.1}
\caption{PolyU \cite{xu1804real}.}
\label{table_3}
\centering
\begin{tabular}{lcccccccc}
  \toprule

Method
& {DnCNN}\cite{7839189}
& {FFDNet}\cite{8365806}	
& {CBDNet}\cite{8954448}
& {RINDNet}\cite{9010324}
& {AINDNet}\cite{9157160}
& {ADNet}\cite{tian2020attention}
& {WMDCNN}\cite{tian2023multi}
& {Restormer}\cite{zamir2022restormer}
\\

Venue
& {\textit{TIP'2017}}
& {\textit{TIP'2018}}
& {\textit{CVPR'2019}}
& {\textit{ICCV'2019}}
& {\textit{CVPR'2020}}
& {\textit{NN'2020}}
& {\textit{PR'2023}}
& {\textit{CVPR'2022}}
\\

PSNR
& 36.08
& 37.19
& 36.93
& 38.57
& 37.22
& 37.02
& 37.07
& 37.66
 \\

SSIM
& 0.916
& 0.939
& 0.946
& 0.970
& 0.946
& 0.939
& 0.941
& 0.956
 \\\hline

Method
& {R2R}\cite{9577798}
& {B2U}\cite{Wang_2022_CVPR}
&{ZS-N2N}\cite{10203304}
&{SDAP}\cite{Pan_2023_ICCV}
&{BM3D}\cite{4378954}
&{WNNM}\cite{6909762}
&{MCWNNM}\cite{8237387}
&{TWSC}\cite{xu2018trilateral}
\\

Venue
& {\textit{CVPR'2021}}
& {\textit{CVPR'2022}}
& {\textit{CVPR'2023}}
& {\textit{ICCV'2023}}
& {\textit{TIP'2007}}
& {\textit{CVPR'2014}}
& {\textit{ICCV'2017}}
& {\textit{ECCV'2018}}

\\

PSNR

& 38.47
& 38.25
& 36.04
& 34.30
& 37.40
& 36.59
& 38.51
& 36.10
 \\

SSIM

& 0.965
& 0.968
& 0.915
& 0.923
& 0.953
& 0.925
& 0.967
& 0.963
 \\\hline

Method

& {DIP}\cite{Ulyanov2017DeepIP}
& {SB}\cite{54}
& {DRP-DIP}\cite{Li_2023_CVPR}
& {NN+BM3D}\cite{zheng2021unsupervised}
& {PD}\cite{zhou2020awgn}
& {ScoreDVI}\cite{cheng2023score}
& {Net2Net}     \\

Venue
& {\textit{IJCV'2020}}
& {\textit{IJCV'2022}}
& {\textit{CVPR'2023}}
& {\textit{ICLR'2021}}
& {\textit{AAAI'2020}}
& {\textit{ICCV'2023}}
\\

PSNR
& 37.17
& 37.11
& 36.10
& 38.74
& 37.04
& 37.77
& \textbf{38.82} \\

SSIM
& 0.912
& 0.911
& 0.886
& 0.968
& 0.940
& 0.959
& \textbf{0.969} \\\bottomrule

\end{tabular}
\end{table*}

\section{Conclusion}\label{7}

To address the problem of overfitting noise in image restoration caused by the failure of the spectral shift mechanism in deep structural priors, this chapter proposes Net2Net, an image restoration algorithm driven by deep structural prior regularization. Net2Net introduces a pretrained deep unbiased denoising prior (DRUNet) as an explicit prior regularization constraint within the data fitting framework of the deep structural prior. Through the RED framework, the deep denoising prior model is co-optimized with the untrained network model in DIP, thereby introducing a stronger external explicit deep prior regularization constraint on top of the original implicit network structural prior in DIP. Net2Net enhances the model's ability to represent degradation priors and effectively suppresses overfitting. Furthermore, a detailed analysis of the feasibility of the Net2Net optimization solution is conducted, verifying that the pretrained deep denoising prior (DRUNet) simultaneously satisfies the three RED conditions of differentiability, local homogeneity, and Jacobian symmetry in the RED solution framework. The authors also analyze how these conditions ensure the validity of the RED gradient, thereby ensuring the efficient solution of the Net2Net model. Finally, the experimental results show that Net2Net can effectively avoid overfitting to noise and show excellent and stable recovery capabilities.

\section*{Acknowledgements}
A preliminary version of this paper was presented in \cite{yuan2024mixed}. Compared with the conference version, we have made significant improvements. Specifically, we have provided a detailed introduction to related works and preliminaries. We have also emphasized the research motivation and offered a more comprehensive discussion of the algorithm details. We have added and compared several methods proposed in recent years. The superiority of our proposed method has been thoroughly validated on newly added datasets. This work was supported in part by the National Natural Science Foundation of China (Grants Nos. 62271016, 92148206) and the Beijing Natural Science Foundation under Grant 4222007.

\bibliographystyle{IEEEtran}
\bibliography{reference}

\begin{thebibliography}{10}
\providecommand{\url}[1]{#1}
\csname url@samestyle\endcsname
\providecommand{\newblock}{\relax}
\providecommand{\bibinfo}[2]{#2}
\providecommand{\BIBentrySTDinterwordspacing}{\spaceskip=0pt\relax}
\providecommand{\BIBentryALTinterwordstretchfactor}{4}
\providecommand{\BIBentryALTinterwordspacing}{\spaceskip=\fontdimen2\font plus
\BIBentryALTinterwordstretchfactor\fontdimen3\font minus \fontdimen4\font\relax}
\providecommand{\BIBforeignlanguage}[2]{{%
\expandafter\ifx\csname l@#1\endcsname\relax
\typeout{** WARNING: IEEEtran.bst: No hyphenation pattern has been}%
\typeout{** loaded for the language `#1'. Using the pattern for}%
\typeout{** the default language instead.}%
\else
\language=\csname l@#1\endcsname
\fi
#2}}
\providecommand{\BIBdecl}{\relax}
\BIBdecl

\bibitem{buades2005review}
A.~Buades, B.~Coll, and J.-M. Morel, ``A review of image denoising algorithms, with a new one,'' \emph{Multiscale modeling \& simulation}, vol.~4, no.~2, pp. 490--530, 2005.

\bibitem{yuan2024simultaneous}
W.~Yuan, Y.~Wang, R.~Fan, Y.~Zhang, G.~Wei, C.~Meng, and X.~Bai, ``Simultaneous image denoising and completion through convolutional sparse representation and nonlocal self-similarity,'' \emph{Computer Vision and Image Understanding}, vol. 249, p. 104216, 2024.

\bibitem{elad2023image}
M.~Elad, B.~Kawar, and G.~Vaksman, ``Image denoising: The deep learning revolution and beyond—a survey paper,'' \emph{SIAM Journal on Imaging Sciences}, vol.~16, no.~3, pp. 1594--1654, 2023.

\bibitem{izadi2023image}
S.~Izadi, D.~Sutton, and G.~Hamarneh, ``Image denoising in the deep learning era,'' \emph{Artificial Intelligence Review}, vol.~56, no.~7, pp. 5929--5974, 2023.

\bibitem{milanfar2024denoising}
P.~Milanfar and M.~Delbracio, ``Denoising: A powerful building-block for imaging, inverse problems, and machine learning,'' \emph{arXiv preprint arXiv:2409.06219}, 2024.

\bibitem{yuan2024weighted}
W.~Yuan, C.~Meng, and X.~Bai, ``Weighted side-window based gradient guided image filtering,'' \emph{Pattern Recognition}, vol. 146, p. 110006, 2024.

\bibitem{yuan2023guided}
W.~Yuan, Y.~Wang, C.~Meng, and X.~Bai, ``Guided image filtering: A survey and evaluation study,'' in \emph{Proceedings of the 5th ACM International Conference on Multimedia in Asia Workshops}, 2023, pp. 1--5.

\bibitem{li2025gs2pose}
J.~Li, W.~Yuan, Y.~Wang, Y.~Zeng, S.~Shu, C.~Meng, and X.~Bai, ``Gs2pose: Marry gaussian splatting to 6d object pose estimation,'' \emph{arXiv preprint arXiv:2510.16777}, 2025.

\bibitem{8026108}
L.~Zhang and W.~Zuo, ``Image restoration: From sparse and low-rank priors to deep priors [lecture notes],'' \emph{IEEE Signal Processing Magazine}, vol.~34, no.~5, pp. 172--179, 2017.

\bibitem{yuan2025image}
W.~Yuan, C.~Meng, and X.~Bai, ``Image restoration driven by dual-scale prior,'' \emph{Neural Networks}, p. 108138, 2025.

\bibitem{wang2025mamba}
Y.~Wang, T.~Guo, W.~Yuan, S.~Shu, C.~Meng, and X.~Bai, ``Mamba-based deformable medical image registration with an annotated brain mr-ct dataset,'' \emph{Computerized Medical Imaging and Graphics}, p. 102566, 2025.

\bibitem{yuan2025guided}
W.~Yuan, Y.~Wang, C.~Meng, and X.~Bai, ``Guided image filtering-conventional to deep models: A review and evaluation study,'' \emph{Computer Vision and Image Understanding}, p. 104278, 2025.

\bibitem{wang2024samihs}
Y.~Wang, K.~Chen, W.~Yuan, Z.~Tang, C.~Meng, and X.~Bai, ``Samihs: adaptation of segment anything model for intracranial hemorrhage segmentation,'' in \emph{2024 IEEE International Symposium on Biomedical Imaging (ISBI)}.\hskip 1em plus 0.5em minus 0.4em\relax IEEE, 2024, pp. 1--5.

\bibitem{li2024single}
Z.~Li, C.~Meng, D.~Li, L.~Liu, W.~Yuan, and Z.~Huang, ``Single view-based pose estimation of unknown circular object with rgb-d camera,'' \emph{IEEE transactions on aerospace and electronic systems}, vol.~60, no.~4, pp. 3954--3966, 2024.

\bibitem{wang2023coam}
Y.~Wang, W.~Yuan, and X.~Bai, ``Coam-net: coordinate asymmetric multi-scale fusion strategy for polyp segmentation: Y. wang et al.'' \emph{Applied Intelligence}, vol.~53, no.~24, pp. 30\,626--30\,641, 2023.

\bibitem{meng2023oscillating}
C.~Meng, D.~Li, W.~Yuan, K.~Wu, and H.~Shen, ``Oscillating saw calibration for mandibular osteotomy robots,'' \emph{Applied Sciences}, vol.~13, no.~17, p. 9773, 2023.

\bibitem{li2022review}
D.~Li, W.~Yuan, Z.~Li, C.~Meng, and Q.~Sun, ``A review of solutions to stereo correspondence challenges,'' in \emph{Chinese Conference on Image and Graphics Technologies}.\hskip 1em plus 0.5em minus 0.4em\relax Springer, 2022, pp. 86--102.

\bibitem{yuan2021efficient}
W.~Yuan, C.~Meng, X.~Tong, and Z.~Li, ``Efficient local stereo matching algorithm based on fast gradient domain guided image filtering,'' \emph{Signal Processing: Image Communication}, vol.~95, p. 116280, 2021.

\bibitem{yuan2019sgm}
W.~Yuan, X.~Tong, and B.~Xiao, ``Sgm-based disparity estimation under radiometric variations,'' in \emph{Chinese Conference on Image and Graphics Technologies}.\hskip 1em plus 0.5em minus 0.4em\relax Springer, 2019, pp. 382--391.

\bibitem{7839189}
K.~Zhang, W.~Zuo, Y.~Chen, D.~Meng, and L.~Zhang, ``Beyond a gaussian denoiser: Residual learning of deep cnn for image denoising,'' \emph{IEEE Trans. Image Process.,}, vol.~26, no.~7, pp. 3142--3155, 2017.

\bibitem{8365806}
K.~Zhang, W.~Zuo, and L.~Zhang, ``Ffdnet: Toward a fast and flexible solution for cnn-based image denoising,'' \emph{IEEE Trans. Image Process.,}, vol.~27, no.~9, pp. 4608--4622, 2018.

\bibitem{Ryou_2024_CVPR}
D.~Ryou, I.~Ha, H.~Yoo, D.~Kim, and B.~Han, ``Robust image denoising through adversarial frequency mixup,'' in \emph{Proceedings of the IEEE/CVF Conference on Computer Vision and Pattern Recognition (CVPR)}, June 2024, pp. 2723--2732.

\bibitem{7780555}
S.~Nam, Y.~Hwang, Y.~Matsushita, and S.~J. Kim, ``A holistic approach to cross-channel image noise modeling and its application to image denoising,'' in \emph{Proc. IEEE Conf. Comput. Vis. Pattern Recognit. (CVPR)}, 2016, pp. 1683--1691.

\bibitem{8578280}
A.~Abdelhamed, S.~Lin, and M.~S. Brown, ``A high-quality denoising dataset for smartphone cameras,'' in \emph{2018 IEEE/CVF Conference on Computer Vision and Pattern Recognition}, 2018, pp. 1692--1700.

\bibitem{8099777}
T.~Plötz and S.~Roth, ``Benchmarking denoising algorithms with real photographs,'' in \emph{2017 IEEE Conference on Computer Vision and Pattern Recognition (CVPR)}, 2017, pp. 2750--2759.

\bibitem{xu1804real}
J.~Xu, H.~Li, Z.~Liang, D.~Zhang, and L.~Zhang, ``Real-world noisy image denoising: A new benchmark. arxiv 2018,'' \emph{arXiv preprint arXiv:1804.02603}.

\bibitem{8953965}
Y.~Zhang, Y.~Zhu, E.~Nichols, Q.~Wang, S.~Zhang, C.~Smith, and S.~Howard, ``A poisson-gaussian denoising dataset with real fluorescence microscopy images,'' in \emph{2019 IEEE/CVF Conference on Computer Vision and Pattern Recognition (CVPR)}, 2019, pp. 11\,702--11\,710.

\bibitem{lebrun2015noise}
M.~Lebrun, M.~Colom, and J.-M. Morel, ``The noise clinic: a blind image denoising algorithm,'' \emph{Image Processing On Line}, vol.~5, pp. 1--54, 2015.

\bibitem{anaya2018renoir}
J.~Anaya and A.~Barbu, ``Renoir--a dataset for real low-light image noise reduction,'' \emph{Journal of Visual Communication and Image Representation}, vol.~51, pp. 144--154, 2018.

\bibitem{8954448}
S.~Guo, Z.~Yan, K.~Zhang, W.~Zuo, and L.~Zhang, ``Toward convolutional blind denoising of real photographs,'' in \emph{Proc. IEEE Conf. Comput. Vis. Pattern Recognit. (CVPR)}, 2019, pp. 1712--1722.

\bibitem{9010324}
S.~Anwar and N.~Barnes, ``Real image denoising with feature attention,'' in \emph{Proc. IEEE/CVF Int. Conf. Comput. Vis. (ICCV)}, 2019, pp. 3155--3164.

\bibitem{9157160}
Y.~Kim, J.~W. Soh, G.~Y. Park, and N.~I. Cho, ``Transfer learning from synthetic to real-noise denoising with adaptive instance normalization,'' in \emph{Proc. IEEE Conf. Comput. Vis. Pattern Recognit. (CVPR)}, 2020, pp. 3479--3489.

\bibitem{9577798}
T.~Pang, H.~Zheng, Y.~Quan, and H.~Ji, ``Recorrupted-to-recorrupted: Unsupervised deep learning for image denoising,'' in \emph{Proc. IEEE Conf. Comput. Vis. Pattern Recognit. (CVPR)}, 2021, pp. 2043--2052.

\bibitem{Wang_2022_CVPR}
Z.~Wang, J.~Liu, G.~Li, and H.~Han, ``Blind2unblind: Self-supervised image denoising with visible blind spots,'' in \emph{Proc. IEEE Conf. Comput. Vis. Pattern Recognit. (CVPR)}, June 2022, pp. 2027--2036.

\bibitem{10203304}
Y.~Mansour and R.~Heckel, ``Zero-shot noise2noise: Efficient image denoising without any data,'' in \emph{Proc. IEEE Conf. Comput. Vis. Pattern Recognit. (CVPR)}, 2023, pp. 14\,018--14\,027.

\bibitem{Pan_2023_ICCV}
Y.~Pan, X.~Liu, X.~Liao, Y.~Cao, and C.~Ren, ``Random sub-samples generation for self-supervised real image denoising,'' in \emph{Proc. IEEE/CVF Int. Conf. Comput. Vis. (ICCV)}, October 2023, pp. 12\,150--12\,159.

\bibitem{xx040605412}
S.~Osher, M.~Burger, D.~Goldfarb, J.~Xu, and W.~Yin, ``An iterative regularization method for total variation-based image restoration,'' \emph{Multiscale Modeling \& Simulation}, vol.~4, no.~2, pp. 460--489, 2005.

\bibitem{6814320}
J.~{Zhang}, D.~{Zhao}, and W.~{Gao}, ``Group-based sparse representation for image restoration,'' \emph{IEEE Trans. Image Process}, vol.~23, no.~8, pp. 3336--3351, Aug. 2014.

\bibitem{6909762}
S.~Gu, L.~Zhang, W.~Zuo, and X.~Feng, ``Weighted nuclear norm minimization with application to image denoising,'' in \emph{Proc. IEEE Conf. Comput. Vis. Pattern Recognit. (CVPR)}, June 2014, pp. 2862--2869.

\bibitem{4378954}
K.~Dabov, A.~Foi, V.~Katkovnik, and K.~Egiazarian, ``Color image denoising via sparse 3d collaborative filtering with grouping constraint in luminance-chrominance space,'' in \emph{2007 IEEE International Conference on Image Processing}, vol.~1, 2007, pp. I -- 313--I -- 316.

\bibitem{kohler2019toward}
T.~K{\"o}hler, M.~B{\"a}tz, F.~Naderi, A.~Kaup, A.~Maier, and C.~Riess, ``Toward bridging the simulated-to-real gap: Benchmarking super-resolution on real data,'' \emph{IEEE transactions on pattern analysis and machine intelligence}, vol.~42, no.~11, pp. 2944--2959, 2019.

\bibitem{zhou2020awgn}
Y.~Zhou, J.~Jiao, H.~Huang, Y.~Wang, J.~Wang, H.~Shi, and T.~Huang, ``When awgn-based denoiser meets real noises,'' in \emph{Proceedings of the AAAI Conference on Artificial Intelligence}, vol.~34, no.~07, 2020, pp. 13\,074--13\,081.

\bibitem{9577357}
Y.~Liu, Z.~Qin, S.~Anwar, P.~Ji, D.~Kim, S.~Caldwell, and T.~Gedeon, ``Invertible denoising network: A light solution for real noise removal,'' in \emph{2021 IEEE/CVF Conference on Computer Vision and Pattern Recognition (CVPR)}, 2021, pp. 13\,360--13\,369.

\bibitem{liu2007automatic}
C.~Liu, R.~Szeliski, S.~B. Kang, C.~L. Zitnick, and W.~T. Freeman, ``Automatic estimation and removal of noise from a single image,'' \emph{IEEE transactions on pattern analysis and machine intelligence}, vol.~30, no.~2, pp. 299--314, 2007.

\bibitem{lehtinen18a}
J.~Lehtinen, J.~Munkberg, J.~Hasselgren, S.~Laine, T.~Karras, M.~Aittala, and T.~Aila, ``{N}oise2{N}oise: Learning image restoration without clean data,'' in \emph{Proceedings of the 35th International Conference on Machine Learning}, 10--15 Jul 2018, pp. 2965--2974.

\bibitem{Neshatavar_2022_CVPR}
R.~Neshatavar, M.~Yavartanoo, S.~Son, and K.~M. Lee, ``Cvf-sid: Cyclic multi-variate function for self-supervised image denoising by disentangling noise from image,'' in \emph{Proceedings of the IEEE/CVF Conference on Computer Vision and Pattern Recognition (CVPR)}, June 2022, pp. 17\,583--17\,591.

\bibitem{9878719}
W.~Lee, S.~Son, and K.~M. Lee, ``Ap-bsn: Self-supervised denoising for real-world images via asymmetric pd and blind-spot network,'' in \emph{2022 IEEE/CVF Conference on Computer Vision and Pattern Recognition (CVPR)}, 2022, pp. 17\,704--17\,713.

\bibitem{chen2024exploring}
S.~Chen, J.~Zhang, Z.~Yu, and T.~Huang, ``Exploring efficient asymmetric blind-spots for self-supervised denoising in real-world scenarios,'' in \emph{Proceedings of the IEEE/CVF Conference on Computer Vision and Pattern Recognition}, 2024, pp. 2814--2823.

\bibitem{Ulyanov2017DeepIP}
D.~Ulyanov, A.~Vedaldi, and V.~S. Lempitsky, ``Deep image prior,'' \emph{International Journal of Computer Vision}, vol. 128, pp. 1867 -- 1888, 2020.

\bibitem{heckel2018deep}
R.~Heckel and P.~Hand, ``Deep decoder: Concise image representations from untrained non-convolutional networks,'' in \emph{Proc. Int. Conf. Learn. Represent. (ICLR)}, 2019.

\bibitem{NAS-DIP}
Y.-C. Chen, C.~Gao, E.~Robb, and J.-B. Huang, ``Nas-dip: Learning deep image prior with neural architecture search,'' in \emph{Proc. Eur. Conf. Comput. Vis. (ECCV)}, 2020, pp. 442--459.

\bibitem{Arican2021ISNASDIPIN}
M.~E. Arican, O.~Kara, G.~Bredell, and E.~Konukoglu, ``Isnas-dip: Image-specific neural architecture search for deep image prior,'' \emph{2022 IEEE/CVF Conference on Computer Vision and Pattern Recognition (CVPR)}, pp. 1950--1958, 2021.

\bibitem{54}
Z.~{Shi}, M.~{Pascal}, and S.~{Subhransu}, ``On measuring and controlling the spectral bias of the deep image prior,'' \emph{IJCV.}, vol. 130, no.~4, pp. 885--908, 2022.

\bibitem{Li_2023_CVPR}
T.~Li, H.~Wang, Z.~Zhuang, and J.~Sun, ``Deep random projector: Accelerated deep image prior,'' in \emph{CVPR}, June 2023, pp. 18\,176--18\,185.

\bibitem{venkatakrishnan2013plug}
S.~V. Venkatakrishnan, C.~A. Bouman, and B.~Wohlberg, ``Plug-and-play priors for model based reconstruction,'' in \emph{2013 IEEE global conference on signal and information processing}.\hskip 1em plus 0.5em minus 0.4em\relax IEEE, 2013, pp. 945--948.

\bibitem{doi:10.1137/16M1102884}
Y.~Romano, M.~Elad, and P.~Milanfar, ``The little engine that could: Regularization by denoising (red),'' \emph{SIAM Journal on Imaging Sciences}, vol.~10, no.~4, pp. 1804--1844, 2017.

\bibitem{cohen2021regularization}
R.~Cohen, M.~Elad, and P.~Milanfar, ``Regularization by denoising via fixed-point projection (red-pro),'' \emph{SIAM Journal on Imaging Sciences}, vol.~14, no.~3, pp. 1374--1406, 2021.

\bibitem{Reehorst2018RegularizationBD}
E.~T. Reehorst and P.~Schniter, ``Regularization by denoising: Clarifications and new interpretations,'' \emph{IEEE TCI}, vol.~5, pp. 52--67, 2018.

\bibitem{56}
S.~Boyd, N.~Parikh, E.~Chu, B.~Peleato, and J.~Eckstein, ``Distributed optimization and statistical learning via the alternating direction method of multipliers,'' \emph{Found. Trends Mach. Learn.}, vol.~3, no.~1, p. 1–122, 2011.

\bibitem{ronneberger2015u}
O.~Ronneberger, P.~Fischer, and T.~Brox, ``U-net: Convolutional networks for biomedical image segmentation,'' in \emph{Proc. Int. Conf. Med. Image Comput. Comput.-Assisted Intervention (MICCAI)}, 2015, pp. 234--241.

\bibitem{he2016deep}
K.~He, X.~Zhang, S.~Ren, and J.~Sun, ``Deep residual learning for image recognition,'' in \emph{Proceedings of the IEEE conference on computer vision and pattern recognition}, 2016, pp. 770--778.

\bibitem{mohan2019robust}
S.~Mohan, Z.~Kadkhodaie, E.~P. Simoncelli, and C.~Fernandez-Granda, ``Robust and interpretable blind image denoising via bias-free convolutional neural networks,'' \emph{arXiv preprint arXiv:1906.05478}, 2019.

\bibitem{1284395}
Z.~Wang, A.~C. Bovik, H.~R. Sheikh, and E.~P. Simoncelli, ``Image quality assessment: from error visibility to structural similarity,'' \emph{IEEE TIP.}, vol.~13, no.~4, pp. 600--612, Apr. 2004.

\bibitem{yuan2024mixed}
W.~Yuan, Y.~Wang, N.~Li, C.~Meng, and X.~Bai, ``Mixed degradation image restoration via deep image prior empowered by deep denoising engine,'' in \emph{2024 International Joint Conference on Neural Networks (IJCNN)}.\hskip 1em plus 0.5em minus 0.4em\relax IEEE, 2024, pp. 1--8.

\bibitem{faye2024regularization}
E.~C. Faye, M.~D. Fall, and N.~Dobigeon, ``Regularization by denoising: Bayesian model and langevin-within-split gibbs sampling,'' \emph{arXiv preprint arXiv:2402.12292}, 2024.

\bibitem{1467423}
A.~{Buades}, B.~{Coll}, and J.~. {Morel}, ``A non-local algorithm for image denoising,'' in \emph{Proc. IEEE Comput. Vis. Pattern Recog.}, vol.~2, June 2005, pp. 60--65 vol. 2.

\bibitem{7527621}
Y.~Chen and T.~Pock, ``Trainable nonlinear reaction diffusion: A flexible framework for fast and effective image restoration,'' \emph{IEEE TPAMI}, vol.~39, no.~6, pp. 1256--1272, 2017.

\bibitem{945431111}
K.~Zhang, Y.~Li, W.~Zuo, L.~Zhang, L.~Van~Gool, and R.~Timofte, ``Plug-and-play image restoration with deep denoiser prior,'' \emph{IEEE Trans. Pattern Anal. Mach. Intell.,}, vol.~44, no.~10, pp. 6360--6376, 2022.

\bibitem{rudin1964principles}
W.~Rudin \emph{et~al.}, \emph{Principles of mathematical analysis}.\hskip 1em plus 0.5em minus 0.4em\relax McGraw-hill New York, 1964, vol.~3.

\bibitem{9157420}
Y.~Quan, M.~Chen, T.~Pang, and H.~Ji, ``Self2self with dropout: Learning self-supervised denoising from single image,'' in \emph{2020 IEEE/CVF Conference on Computer Vision and Pattern Recognition (CVPR)}, 2020, pp. 1887--1895.

\bibitem{zheng2021unsupervised}
D.~Zheng, S.~H. Tan, X.~Zhang, Z.~Shi, K.~Ma, and C.~Bao, ``An unsupervised deep learning approach for real-world image denoising,'' in \emph{International Conference on Learning Representations}, 2021.

\bibitem{cheng2023score}
J.~Cheng, T.~Liu, and S.~Tan, ``Score priors guided deep variational inference for unsupervised real-world single image denoising,'' in \emph{Proceedings of the IEEE/CVF International Conference on Computer Vision}, 2023, pp. 12\,937--12\,948.

\bibitem{tian2020attention}
C.~Tian, Y.~Xu, Z.~Li, W.~Zuo, L.~Fei, and H.~Liu, ``Attention-guided cnn for image denoising,'' \emph{Neural Networks}, vol. 124, pp. 117--129, 2020.

\bibitem{tian2023multi}
C.~Tian, M.~Zheng, W.~Zuo, B.~Zhang, Y.~Zhang, and D.~Zhang, ``Multi-stage image denoising with the wavelet transform,'' \emph{Pattern Recognit.}, vol. 134, p. 109050, 2023.

\bibitem{zamir2022restormer}
S.~W. Zamir, A.~Arora, S.~Khan, M.~Hayat, F.~S. Khan, and M.-H. Yang, ``Restormer: Efficient transformer for high-resolution image restoration,'' in \emph{Proc. IEEE Conf. Comput. Vis. Pattern Recognit. (CVPR)}, 2022, pp. 5728--5739.

\bibitem{8237387}
J.~Xu, L.~Zhang, D.~Zhang, and X.~Feng, ``Multi-channel weighted nuclear norm minimization for real color image denoising,'' in \emph{2017 IEEE International Conference on Computer Vision (ICCV)}, 2017, pp. 1105--1113.

\bibitem{xu2018trilateral}
J.~Xu, L.~Zhang, and D.~Zhang, ``A trilateral weighted sparse coding scheme for real-world image denoising,'' in \emph{Proceedings of the European conference on computer vision (ECCV)}, 2018, pp. 20--36.

\end{thebibliography}

\end{document}